\documentclass[letterpaper, 10pt, conference]{ieeeconf}      

\IEEEoverridecommandlockouts                              

\usepackage{graphics} 

\usepackage{graphicx}
\graphicspath{ {./graphics/}{../} }

\usepackage{xcolor}
\usepackage{subcaption}
\newcommand{\ph}[1]{{\textbf{#1}:}} 
\newcommand{\comment}[1]{} 

\usepackage[normalem]{ulem}
\usepackage{url}
\usepackage{amsmath}
\usepackage{amsfonts}
\usepackage{amssymb}
\usepackage{soul}
\usepackage{textcomp}
\let\labelindent\relax
\usepackage{enumitem}

\usepackage{hyperref}

\title{\LARGE \bf
Autonomous Spot: \\
Long-Range Autonomous Exploration of \\Extreme Environments with Legged Locomotion
(preprint version)
}

\author{Amanda Bouman$^{*2}$, Muhammad Fadhil Ginting$^{*1}$, Nikhilesh Alatur$^{*2}$, Matteo Palieri$^{1}$, \\
David D. Fan$^{1}$, Thomas Touma$^{2}$, Torkom Pailevanian$^{1}$, Sung-Kyun Kim$^{1}$, Kyohei Otsu$^{1}$,\\
Joel Burdick$^{2}$, and Ali-akbar Agha-mohammadi$^{1}$
\thanks{$^*$These authors contributed equally to this work.}
\thanks{$^{1}$NASA Jet Propulsion Laboratory, California Institute of Technology, Pasadena, CA, USA,
     {\tt\small \{firstname.lastname\}@jpl.nasa.gov}}%
\thanks{$^{2}$Department of Mechanical and Civil Engineering, Division of Engineering and Applied Science, California Institute of Technology, Pasadena, CA, USA,
         {\tt\small \{abouman,jwb\}@caltech.edu}}%
}

\usepackage[absolute,showboxes]{textpos}
\TPMargin{5pt}
\newcommand{\copyrightstatement}{
    \tekstblokroolkulur{white}%
    \begin{textblock}{0.84}(0.08,0.93)    
         \noindent
         \footnotesize
         \centering
         \copyright 2020. All rights reserved.
    \end{textblock}
}

\begin{document}
\copyrightstatement  
\maketitle
\thispagestyle{empty}
\pagestyle{empty}
\begin{abstract}
This paper serves as one of the first efforts to enable large-scale and long-duration autonomy using the Boston Dynamics Spot robot. Motivated by exploring extreme environments, particularly those involved in the DARPA Subterranean Challenge, this paper pushes the boundaries of the state-of-practice in enabling legged robotic systems to accomplish real-world complex missions in relevant scenarios. 
In particular, we discuss the behaviors and capabilities which emerge from the integration of the autonomy architecture NeBula (Networked Belief-aware Perceptual Autonomy) with next-generation mobility systems.
We will discuss the hardware and software challenges, and solutions in mobility, perception, autonomy, and very briefly, wireless networking, as well as lessons learned and future directions. 
We demonstrate the performance of the proposed solutions on physical systems in real-world scenarios.\footnote[3]{Selected mission clips: \url{https://youtu.be/MSKEaPtYOLI}}
The proposed solution contributed to winning 1st-place in the 2020 DARPA Subterranean Challenge, Urban Circuit.\footnote[4]{CoSTAR Team Website: \url{https://costar.jpl.nasa.gov}}
\end{abstract}





\section{Introduction}\label{sec:intro}

Autonomous robot mapping and traversal of extreme environments under time constraints has a wide variety of real-world applications, including search and rescue after natural disasters \cite{Nagatani2013}, exploration of extreme planetary terrains \cite{ali-agu,Taka2020JFR,Husain2013}, and inspection of urban underground environments \cite{Kolvenbach2020}. As a concrete mission, we focus on the DARPA Subterranean (SubT) Challenge \cite{subt_webpage}: a robotic competition 
that targets missions to explore, map, and search extreme underground environments.




Extreme terrains typically involve mobility-stressing elements that can impose conflicting requirements on the development of mobility systems. For example, in the context of the SubT challenge, the systems need to
1) be small enough to move through passages as narrow as 80 cm in diameter while carrying a large-enough payload capable of providing high-levels of sensing, autonomy, computing, and communication capabilities, and 2) remain operational for long-duration missions ($\geq$ 1 hour) while actively exploring large areas (multi-kilometer in length) that require traversal of mobility-stressing features, such as stairs, uneven terrain, and risky, obstacle-laden areas.

\begin{figure}[t!]
  \centering
  \includegraphics[width=0.48\textwidth,trim={0.5cm 0 0.5cm 0},clip]{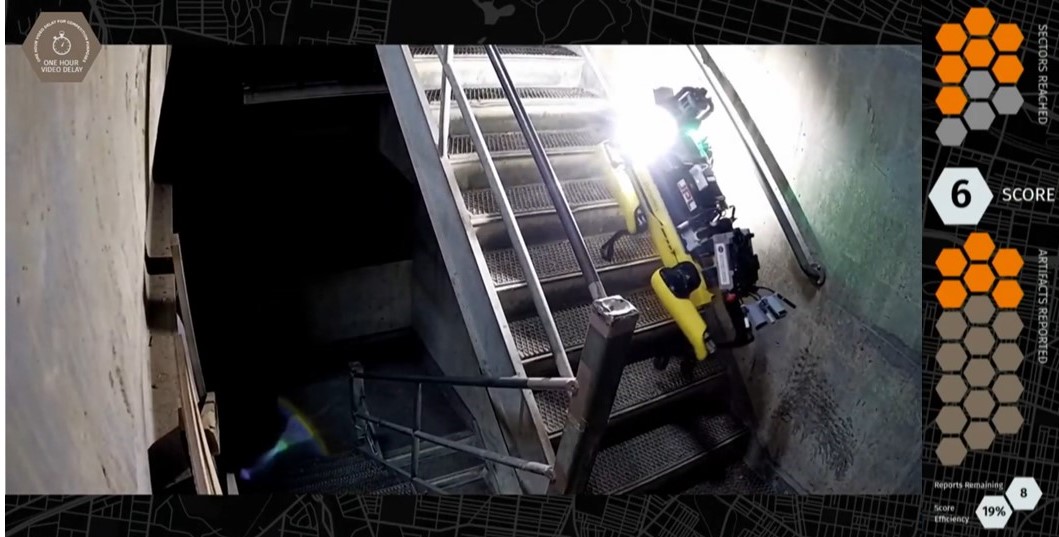}
  \caption{
  Live-mission image of Autonomous Spot robot climbing down four flights of stairs in the Urban Circuit of the DARPA Subterranean Challenge. This platform is one of the elements in team CoSTAR's solution that won the Urban Circuit of this competition. (Image credit: DARPA).} 
  
  
  \label{fig:stairs-firstPage}
\end{figure}

\begin{figure*}[h!]
  \centering
  \includegraphics[width=0.9\textwidth]{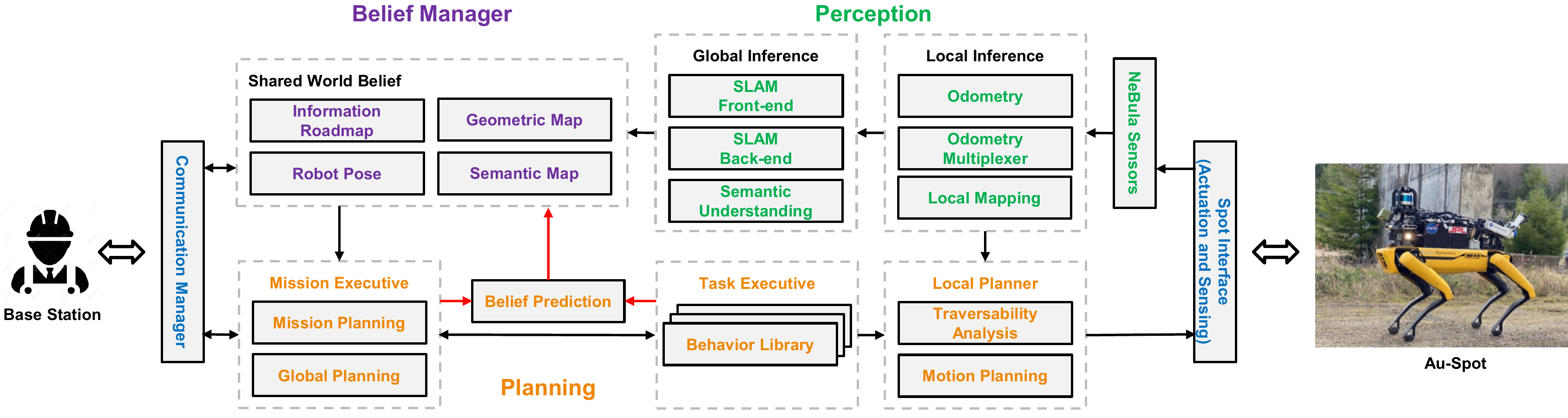}
  \caption{Overview of the NeBula system architecture enabling high-level autonomy on Spot. Red paths denote NeBula's belief-aware planning where the planner aims to minimize mission risk by generating information-seeking paths that increase the accuracy of the shared world belief. 
}
  \label{fig:spot_sysarch}
\end{figure*}

Legged robots offer unique mobility capabilities which make them highly suitable for traversing challenging environments that would prove difficult for wheeled robots, as they have the ability to meet locomotion, size, payload, and endurance requirements to operate in extreme environments. 
For some prominent examples, see: ANYmal~\cite{whyrobotdeepmines,Bellicoso2018}, Robosimian~\cite{Karumanchi2017}, DRC-HUBO+~\cite{jung2018development}, Nimbro Momaro~\cite{schwarz2017nimbro}, MIT Cheetah~\cite{mit_cheetah}, BigDog~\cite{bigdog}, Ghost Robotics Vision 60~\cite{miller2019tunnel}. 


The robotics research community is now in the early stages of empowering legged robots with high levels of autonomy to carry out complex missions in challenging, real-life environments~\cite{Delmerico2019}. Ramezani et al.~\cite{Ramezani2020} equipped the ANYmal quadruped with a LiDAR SLAM framework for autonomous mapping capabilities. The solution in \cite{Ramezani2020} requires manual teleoperation to build an initial map of the environment, upon which the robot can autonomously navigate within the constructed map. The method is demonstrated in an industrial complex.

Bayer et al.~\cite{Bayer2019} demonstrated fully autonomous exploration in rough, single-level, indoor and outdoor terrains.
The researchers augmented an experimental hexapod platform with commercial vision sensors which were used for localization and terrain mapping.
Miller et al.~\cite{miller2019tunnel} 
endowed a Ghost Vision 60 quadruped with higher levels of autonomy to explore a tunnel environment during the 2019 DARPA Subterranean Challenge, Tunnel Circuit. They present one of the first efforts in autonomous legged exploration of an unknown, GPS-denied subterranean environments, focused on single-level, tunnel-like environments. 

\ph{Contributions} 
In this work, we focus on Boston Dynamics' Spot robot as our base mobility platform (Fig. \ref{fig:stairs-firstPage}). 
We briefly discuss the NeBula (Networked Belief-aware Perceptual Autonomy) architecture and explain some of the key elements of integrating comprehensive autonomy 
with the Spot robot.
We describe the behaviors and overall performance of the system in a complex, autonomous mission during the Urban Circuit of the DARPA Subterranean Challenge.
While the main objective of this paper is to provide a system-level overview of the entire autonomy stack, we will describe in deeper detail some specific aspects of the algorithms that are critical to enabling legged autonomy in complex missions. 

Highlights of this paper or areas where we advance the current state-of-practice on Spot and legged robots are:
\begin{enumerate}
    \item Endowing a legged platform with high-level autonomy so that it may traverse kilometer-scale distances in a multi-level, underground, GPS-denied environment within 60 minutes. 
    \item Enabling reliable multi-sensor odometry in perceptually-degraded environments.
    \item Demonstrating perception- and traversability-aware local planning on legged platforms to negotiate challenging terrains 
    and perceptually-degraded environments.
    \item Developing a rugged and lightweight hardware system to equip Spot with the NeBula autonomy package. 
\end{enumerate}
The performance of these technologies was successfully field-tested at the Urban Circuit of the DARPA SubT Challenge (and practice runs leading to the competition), as part of team CoSTAR's solution.

\ph{Outline}
In Section \ref{sec:nebula}, we provide an overview of the NeBula architecture and describe its elements.
In Section \ref{sec:spot}, we discuss the legged mobility system and the hardware payload. Sections \ref{sec:state_estimation}, \ref{sec:local_planning}, and \ref{sec:coverage} focus on selected algorithmic aspects of legged robot odometry, local planning, and high-level mission planning.
Experimental results are presented in Section \ref{sec:experiments}, followed by future work discussion and conclusions.

\section{NeBula Autonomy}\label{sec:nebula}
Motivated by autonomous exploration of extreme surfaces and subsurface terrains on the Moon, Mars and other planetary bodies, NASA's Jet Propulsion Laboratory (NASA JPL) is developing an autonomy architecture referred to as NeBula (Networked Belief-aware Perceptual Autonomy). 
The main focus of NeBula is to provide computationally tractable methods to predict and assess various outcomes and risks in uncertain settings. These methods subsequently enable reliable, coordinated multi-robot exploration of unknown and hard-to-access terrains. To deal with uncertainty in unknown environments, NeBula employs a probabilistic approach. It takes the uncertainty into account to probabilistically fuse various sensing modalities, creates a probabilistic representation of the robot's knowledge of the environment, computes risk, and ``proactively" plans to minimize the mission risk. 



\ph{Architecture} 
Figure \ref{fig:spot_sysarch} illustrates a high-level overview of the NeBula architecture and how its modules are interconnected. 
Spot interface module, which includes Spot's internal locomotion system and inbuilt factory sensors, and NeBula's sensors will be discussed further in Section \ref{sec:spot}.
The odometry module, responsible for measuring and estimating the state and relative motion of the robot, will be discussed in Section \ref{sec:state_estimation}. 
The belief manager block constructs and maintains the robot's model of the environment. 
The planning blocks include the 1) mission planning module that switches between various behaviors such as exploration, stair-climbing, communication-recovery, etc., 2) global planning which guides the coverage behavior and 3) traversability analysis and local motion planning. We will briefly discuss the planning modules in Sections \ref{sec:local_planning} and \ref{sec:coverage}. The communication block is responsible for enabling data exchange between multiple robots and a base station (described in~\cite{Otsu2020}). 

The belief prediction module is a critical component in the NeBula architecture that enables perception-aware and uncertainty-aware planning. This module allows the planner to take perceptual capability into account and helps reduce the risk by increasing the accuracy of the world representation (red arrows in Fig. \ref{fig:spot_sysarch}). We will discuss this feature of NeBula further throughout the paper.
\section{Au-Spot Mobility System}\label{sec:spot}
\ph{Locomotion System} 
Spot is a quadrupedal robot developed by Boston Dynamics to provide mobility on challenging terrain, which may not be negotiated by traditional wheeled robots, including steps and stairs, among others.




\ph{Sensing system} 
Spot's factory perception package from Boston Dynamics comprises five custom RealSenses distributed around the robot. 
To enable higher levels of autonomy required in the SubT challenge, we augment Spot's in-built sensing package with NeBula Sensor Package (NSP). 
The NSP includes a LiDAR, Intel RealSense cameras, high-intensity LEDs, an IMU, gas and wifi detectors, and a thermal camera.
These sensors are integrated into a shock-absorbing, rigid mechanical super structure. 
The NSP can experience significant forces, moments, and vibrations as Spot negotiates complex terrain at high speeds.
A combination of hard resin urethane, semi rigid carbon-infused nylon, and aluminum are used in the manufacturing process for increased structural integrity, sensor data fidelity and lightweight build. 
Further, the design takes into consideration atypical load paths for shock absorption during falls. 

\begin{figure}[t]
  \centering
  \includegraphics[width=0.45\textwidth]{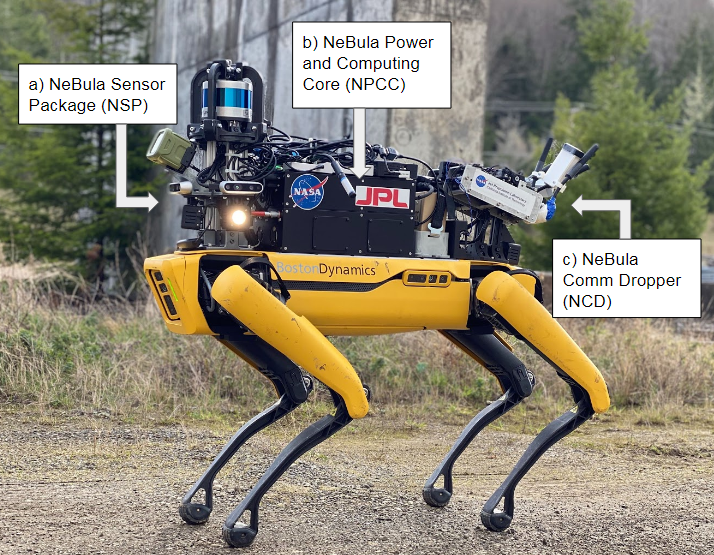}
  \caption{``Autonomous Spot:" Spot powered by NeBula (Au-Spot).}
  \label{fig:spot_full_nebula_form}
\end{figure}

\ph{Power and Computing}
The NeBula Power and Computing Core (NPCC) is designed to mount onto Spot as an auxiliary payload which provides power to all NeBula sensors and computers used for autonomy. The payload enclosure is designed with aluminum to provide protection to the internal electronics if Spot were to fall. 
The payload is powered from an external lithium high capacity battery to provide isolation and extended battery life for Spot's internal battery. 
The NPCC also features a custom power distribution and safety module, which provides fuses, overcurrent protection, overvoltage protection, inrush current limiting and power sequencing of five high efficiency voltage regulators for the sensors, lights, and computers. 
The payload uses two high-power computers 
for sensing, autonomy, and semantic scene understanding.
For brevity, in the rest of this paper, we refer to the combined NSP, NPCC, and Spot robot as Autonomous-Spot or Au-Spot (Fig. \ref{fig:spot_full_nebula_form}).

\section{NeBula Odometry on Legged Systems}\label{sec:state_estimation}
To enable autonomous robotic operation in extreme environments, a reliable odometry source is a prerequisite. In such scenarios, 
darkness, presence of obscurants 
(e.g. dust, fog, smoke), self-similar areas, and  strong platform vibrations caused by mobility-stressing terrains are common features which pose severe challenges to robotic perception.  As a result, accurate odometry estimation on legged platforms is a critical 
challenge, especially under perceptually-degraded conditions.

Uneven and slippery areas make inertial sensing inaccurate while the material composition of the surface where the legged robot is walking on (e.g soft moquette, hard concrete) has strong impacts on the accuracy of kinematic-based odometry (KO). Darkness, or sudden excessive change in illumination, along with dust and the occasional presence of fog and gas, pose significant challenges to cameras. Potential visual aliasing phenomena in texture-less or texture-repetitive environments make feature-tracking problematic, decreasing the overall reliability of vision-based odometry (VO). Self-similar environments with repetitive geometry and lack of distinctive landmarks make scan-matching based methods ambiguous and prone to drift: moreover, active stereo cameras (including the in-built factory ones on the Spot platform) have a limited field of view, which renders them insufficient for our long-range perception applications.

\begin{figure}[t]
  \centering
  \includegraphics[width=0.48\textwidth]{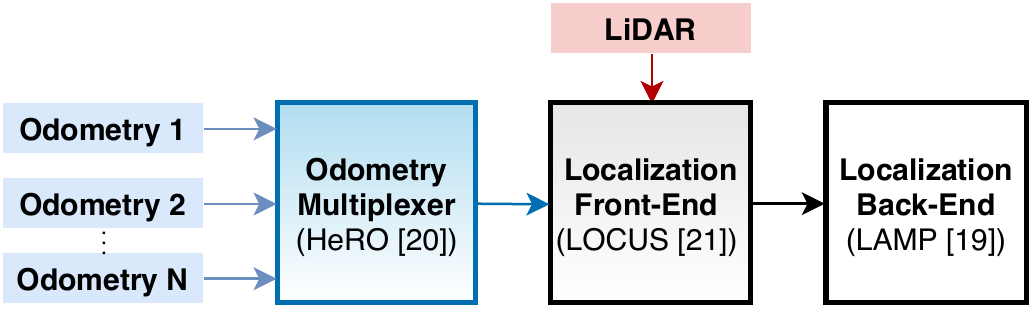}
  \caption{Architecture of the NeBula Multi-Sensor Fusion Framework}
  \label{architecture}
\end{figure}

\ph{Solution Architecture} To overcome these challenges, NeBula relies on a LiDAR-centric uncertainty-aware, multi-sensor fusion framework where a \emph{selected} odometry source is fused as a prior with LiDAR information to enable accurate ego-motion estimation under challenging perceptual conditions. The main components of the proposed approach are: \textit{(i)} an anomaly-aware odometry multiplexer (HeRO), \textit{(ii)} a multi-sensor LiDAR-centric SLAM front-end (LOCUS) and \textit{(iii)} a SLAM back-end (LAMP)~\cite{Ebadi2020}. Fig.~\ref{architecture} provides a high-level overview of the proposed approach. We discuss each component in the following.

\ph{Odometry Multiplexer} To select the best odometry prior to be fused with LiDAR information, we feed multiple and heterogeneous sources of odometry available onboard (e.g., KO, VO, etc.) into an anomaly-aware odometry multiplexer, referred to as HeRO~\cite{Santamaria-navarro2019}. At every time step, HeRO runs a confidence test on each odometry stream (prior) to detect potential anomalies 
(e.g., gaps, jumps, divergences) and identify the most reliable input $\textbf{Y} \in SE(3)$ to be used as a prior in the LiDAR-based front-end.

\ph{Localization Front-End} The output of the odometry multiplexer is fed into a multi-sensor LiDAR-centric SLAM front-end module, referred to as LOCUS~\cite{Palieri2020} that performs a cascaded GICP-based scan-to-scan and scan-to-submap matching operation to estimate the relative motion of the robot between consecutive LiDAR acquisitions. Let $L_{k}$ denote the LiDAR scan acquired the $k$-th time step. We indicate with $\textbf{E}^{k-1}_{k} = \textbf{Y}^{-1}_{k-1}\textbf{Y}_k\in SE(3)$ the rigid body transformation of HeRO's output between two consecutive LiDAR acquisitions. 

In the scan-to-scan matching stage, GICP computes the optimal transformation $\hat{\textbf{T}}^{k-1}_{k}$ that minimizes the residual error $\mathcal{E}$ between corresponding points in $L_{k-1}$ and $L_{k}$. 
\begin{align}\label{eq:GICP_ss}
\hat{\textbf{T}}^{k-1}_{k} = \arg\min_{\textbf{T}^{k-1}_{k}} \mathcal{E} (\textbf{T}^{k-1}_{k}L_{k}, L_{k-1})
\end{align}
To enhance accuracy we initialize the optimization with $\textbf{E}^{k-1}_{k}$. In the case where no input is received by HeRO, we rely on the identity transformation as the prior and the system reverts to pure LiDAR odometry.

To enable global consistency across the history of scans, the motion estimated in the scan-to-scan matching stage is further refined by a scan-to-submap matching step. 
Here, $L_k$ is matched to a local submap $S_k$ which is a robot-centered subset of the global map.  The global map is composed of accumulated past point clouds aligned to the robot pose in the world frame.

\begin{align}\label{eq:GICP_map}
\tilde{\textbf{T}}^{k-1}_{k} = \arg\min_{\textbf{T}^{k-1}_{k}} \mathcal{E} (\textbf{T}^{k-1}_{k}L_{k}, S_{k})
\end{align}
The initial guess of this optimization is $\hat{\textbf{T}}^{k-1}_{k}$, which results from Eq. (\ref{eq:GICP_ss}).
After scan-to-scan and scan-to-submap matching, the final estimated motion $\tilde{\textbf{T}}^{k-1}_{k}$ between consecutive LiDAR acquisitions is used to update the robot pose in the world.


\ph{Localization Back-End} The odometry produced by the front-end is fed into the back-end of our SLAM system, referred to as LAMP~\cite{Ebadi2020} which receives pose-to-pose constraints and solves a Pose Graph Optimization (PGO) and Incremental Consistency Measurement (ICM) problem for global localization when loop closures are detected during traversal.

\section{Local Planning}\label{sec:local_planning}
This section describes our approach to enable Spot to traverse challenging environments, where assessing the traversability risk and planning safe paths is a prerequisite for autonomous navigation.




\subsection{Traversability Map}
We model the traversability of the environment as a grid map $g=(m^1,\cdots,m^n)$ with $n = n_l \times n_w$ cells, arranged in a \textit{locally} 2D grid, where $m^i\in \{safe,lethal\}$ is a Bernoulli random variable.  $m^i=safe$ represents the event that the robot can safely navigate through the $i$-th cell. 
We infer the probability distribution $p(g)$ over grid $g$ and store it as a traversability map.
The steps involved in computing $p(g)$ are detailed in this subsection.




\ph{Risk Sources}
There is a variety of traversability-stressing elements which increase the probability of failure during traversal.
These elements can be purely geometric (e.g., positive obstacles, negative obstacles, steep slopes) or of semantic nature (mission items such as deployed communication nodes or other robots).


\ph{Multi-Fidelity Terrain Map}
For detecting the aforementioned traversability-stressing elements in unknown environments, we build a local terrain map of the robot's surrounding, via data collected from NSP. 
Specifically, we build a multi-fidelity map at various ranges. We use depth cameras for short-range sensing, 
instantaneous LiDAR point clouds for medium-range sensing, 
and spatially fused point clouds \cite{oleynikova2017voxblox} for long-range detection. 
The combination of these various sensing capacities yields an efficient trade-off among \emph{range}, \emph{density} and \emph{accuracy} in the resulting terrain map.

\begin{figure}[t]
  \centering
  \includegraphics[width=\linewidth]{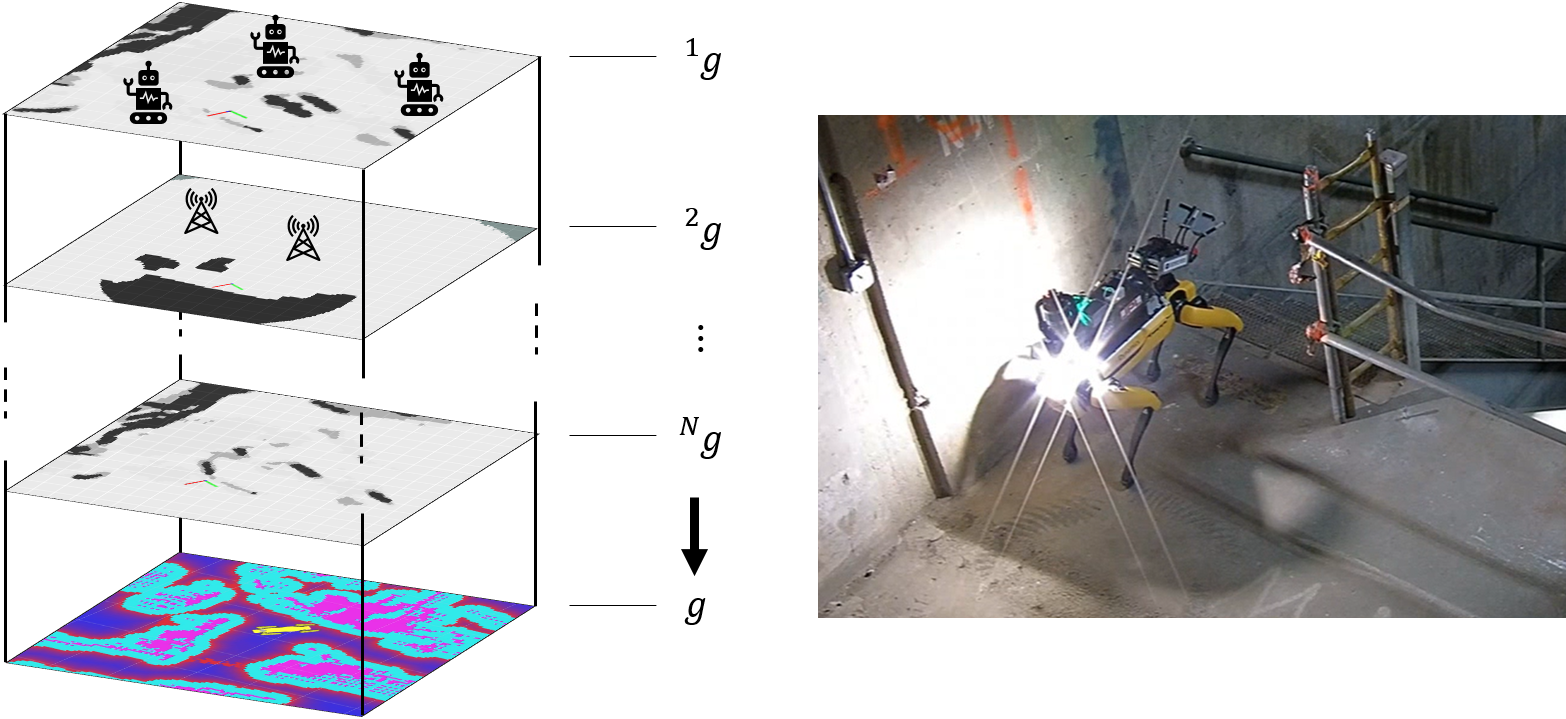}
  \caption{The multi-layer traversability map (${}^{1}g,\cdots,{}^{N}g$), which includes information about various traversability-stressing elements (including other robots in ${}^1g$, communication nodes in ${}^2g$, large negative and positive obstacles in ${}^Ng$, etc.). The bottom map illustrates the aggregated traversability map $g$.
  }

  \label{fig:layered_costmap}
\end{figure}

\ph{Multi-Layer Traversability Map (MLT)}
To capture various traversability-stressing elements, we construct a layered representation (Fig. \ref{fig:layered_costmap}) of the traversability map $g=\{{}^{1}g,...,{}^{N}g\}$, where each layer ${}^{\ell}g$ captures a certain risk source (as explained above). 
At each layer ${\ell}$, we apply algorithms and perform an analysis relevant to the corresponding risk element. 
Specifically, positive and negative obstacles as well as steep slopes are detected on the dense short-range by applying a step filter relative to the local ground plane.
On the medium- and long-range: 
\textit{(i)} Positive obstacles are detected by performing ground segmentation \cite{himmelsbach2010fast} and settling-based collision checks \cite{krusi2017driving}, \textit{(ii)} Negative obstacles by searching for surface discontinuities (holes) in the LiDAR point cloud, and \textit{(iii)} steep slopes by using settling methods such as \cite{krusi2017driving}. Mission items (e.g., deployed communication nodes and other robots) can be detected semantically and superposed on the traversability map.



\ph{Fast Traversability Evaluation}
To enable online generation of MLT as the robot moves, we perform traversability evaluation only on a representative and sparse set of sampled cells $Q = \{i_s\}_{s=0}^S$ in the grid map. 
On each sampled cell $i\in Q$, the traversability risks $p({}^{\ell}m^{i}=lethal)$ are computed and stored in all layers $\ell \in N$. To compute the traversability risk at every cell in the grid, 
we approximate the MLT at a higher resolution by interpolating with a Gaussian kernel or by adding inflation with a radial decay to each query cell $i \in Q$ along the spatial dimension.
Thus, we compute the traversability $p({}^{\ell}m^{n}=safe)$ for all cells $n$ in all layers $\ell$.



\ph{Superposition}
For action generation, we create a single traversability map $g$ by fusing the $N$ different layers $\{{}^{1}g,...,{}^{N}g\}$. 
The information in these layers are not independent in general. We approximate a conservative risk estimate by element-wise multiplication of traversability probabilities:
\begin{align}
p(m^i=safe) = \prod_{\ell=1}^Np({}^{\ell}m^{i}=safe)\hspace{5pt} \forall i = 1,\cdots,n
\end{align}


The bottom layer in Fig. \ref{fig:layered_costmap} illustrates an example traversability map (plotting the maximum likelihood estimation of $p(g)$), obtained during a real mission.



\subsection{Uncertainty and Perception-aware Planning} 
To enable traversability through narrow passages and obstacle-laden environments, one needs to reduce the map uncertainty. We address this challenge by taking into account uncertainty which comes from noisy, asymmetric, and limited sensors in order to find trajectories with minimal path length that also reduce uncertainty in the map $p(g)$ so as to increase the probability that the path taken will be safe. In the NeBula architecture, we formalize this notion with an uncertainty-aware planner.  This general framework can be used to create behaviors which intelligently reduce risk coming from uncertainties in sensing and the environment.  We outline the general framework here.

\ph{Uncertainty-aware Representation}
Let $\mu^i$ be the mean of the Bernoulli distribution of $m^i$.  Then we can model the distribution (or our belief) of $\mu^i$ with its parameters, $p(\mu^i|\hat{\mu}^i,\sigma^{i})$ (e.g. a mean $\hat{\mu}^i$ and variance $\sigma^i$ in the case of a beta distribution). The ``confidence" about $\mu^i$ is captured in $\sigma^i$, where fully unknown and fully known cells have the highest and lowest $\sigma^i$ values, respectively \cite{CRM}. 

\ph{Map prediction}
The uncertainty-aware representation allows us to incorporate perceptual capabilities 
into the planning. We define a policy $\pi$ that returns an ordered sequence of grid locations that the robot visits, along with the orientations of the robot at those locations: $\pi(\cdot)=\{i_k, \theta_k\}_{k=0}^K$. Given the sensors available on the robot and their configuration and noise characteristics, along with a given policy $\pi$, we update our belief of the traversability values in a recursive manner, which we call $\tau$:
\begin{align}
 (\hat{\mu}^i_k, \sigma^i_k) =\tau(\hat{\mu}^i_{k-1}, \sigma^i_{k-1}, z_{k}(\pi) )
\end{align}
where the measurement $z_k(\pi)$ is predicted from a generative model, at the $k$-th time step along the trajectory $\pi$. This becomes increasingly important when the sensor configuration is highly asymmetric on a robot, which is the case for Spot as it has blind spots and areas where sensory measurement noise is considerably higher than other areas.
Maintaining separate probability distributions for individual cells in the map, we predict the map $p(g)$ for $k$-th time step into the future as follows: 
\begin{align}
 g_k \equiv \{(\hat{\mu}^1_k,\sigma^1_k),\cdots,(\hat{\mu}^n_k,\sigma^n_k)\}   
\end{align}

\ph{Risk-Aware/Perception-Aware Planning}
Next we define a risk measure that takes perceptual capabilities and uncertainties into account when planning trajectories.  We also define an optimal policy $\pi^*$ as the policy which minimizes the total path risk $\mathcal{R}^\pi$ along the $K$-step path while moving towards a given goal:
\begin{align}
 \mathcal{R}^{\pi} &= 1 - \prod_{k=0}^K p(m^{i_k}=safe | \hat{\mu}_k^{i},\sigma_k^{i},\pi)
 \label{eq:risk}\\
 \pi^* &= \arg\min_{\pi\in\Pi}\mathbb{E}[\mathcal{R}^{\pi}]
 \label{eq:policy}
\end{align}

Efficient methods for computing predicted risk uncertainty over a 2-D grid for a given sensor model have been considered in~\cite{heiden2017planning}.
When computational complexity is a challenge (with respect to the robot speed), to  enable  efficient  real-time  computation  of  Eq.  (7),  we  rely  on  a  cascaded  policy,  where one can optimize for position and orientation (along the path) sequentially.

\ph{Execution}
We execute the planning problem outlined in Eq. \ref{eq:policy} in a receding-horizon control (RHC) fashion: We optimize for a $K$-step policy. Then, when executing the policy, we select a waypoint at a distance $d$ along the path $\pi^*$, send it to the robot, and while robot is moving towards the waypoint, we resolve the path planning problem to generate a new path from the new robot position. Selecting $d$ is a trade-off between smoothness and tracking error, where a larger $d$ improves stability and smoothness, while a smaller $d$ keeps tracking error of $\pi^*$ lower. The combined effect of perception-aware planning and RHC-based execution will cause Au-Spot to prefer moving in directions that lead to richer sensory input, which leads to a larger reduction in uncertainty via collecting more task-relevant information.

\section{Area coverage and search behavior}\label{sec:coverage}
Our mission planning objective 
is to coordinate a team of autonomous robots to rapidly map and navigate a large (multi-Km), unknown environment characterized by complex topology and terrain under a one-hour time constraint. The global planner realizes this objective by maximizing the area covered by the sensors' footprint along the planned trajectory, within the mission time limit. The planner relies on a representation of the environment with an information-rich graph structure which reduces the policy search space to one that is tractable when exploring large environments over long time horizons.

\ph{Global Planner} We employ a sparse bidirectional graph $G = (V, E)$ that captures the connectivity of the free space in the environment (e.g., \cite{Ali14-IJRR}). A node $v_i \in V$ represents a robot pose, and an edge $e_{ij} \in E$ represents a robot transition between connected nodes $v_i$ and $v_j$. Each node $v_i$ has attached to it a feature vector containing the probability $p_{c}(v_i)$ that the robot has seen a sensor-defined neighborhood around the node. Likewise, each edge $e_{ij}$ will induce a local path computed according to Eq.~\ref{eq:policy}. Then, to each edge $e_{ij}$ we attach a feature vector containing the path length $\ell_{e_{ij}}$ and path traversability risk $\mathcal{R}^{e_{ij}}$ computed by Eq. \ref{eq:risk}.


\ph{Graph Construction} 
We partition graph nodes into frontier nodes $v^f \in V$ and breadcrumb nodes $v^b \in V$. Frontier nodes $v^f$ indicate the areas of the map that have not been fully explored yet, typically, at the boundary of the known and unknown free spaces. A breadcrumb node $v^b$ indicates the areas of the map that have already been fully explored. As the robot moves in the environment to cover and search the space, the graph is expanded through the addition of new breadcrumb and frontier nodes. In other words, visiting a frontier node $v^f$ is expected to lead to changes in the map belief $p(v^f)$, whereas visiting a breadcrumb node will not significantly impact our knowledge of the map coverage.

\ph{Graph Policy}
A graph policy $\lambda$ guides the robot towards the next best node on the graph to maximize the covered area. Specifically, we compute a mapping (feedback policy) $\lambda:V\rightarrow V$ on the graph which maximizes a reward function that encodes a trade-off between coverage information and traversal cost. A macro action $\lambda(v_i)$ induces traversal along edge $e_{ij}$ and updates the graph from $V$ to $V^\prime$. In the following, we discuss steps involved in computing $\lambda$.



\ph{Coverage Information} The \textit{coverage belief} of a graph composed of nodes $\{v_i, \cdots ,v_N\} \in V$ is defined as: 
\begin{align}
    \nonumber
    P_c(V) = \{p_c(v_i), \cdots , p_c(v_N)\}
\end{align}
where $p_c(v_i)$ is the occupancy Bernoulli distribution over a local map centered at node $v_i$. We use entropy to measure the posterior uncertainty of the graph coverage. Entropy of a random variable $x\sim p(x)$ is defined as $H_p(x) = E[-\log p(x)]$. Thus, the graph coverage entropy can be represented as:
\begin{align}
    \nonumber
    H_{p_{c}}\!(V) \!=\! -\!\!\!\sum_{v_i \in V}\!\!
    p_{c}(v_i) \log p_{c}(v_i) 
    \!+\!
    \left(1-p_{c}(v_i)) \! \log (1-p_{c}(v_i)\right)
\end{align}

\ph{Coverage Information Gain}
The coverage information gain (i.e., coverage uncertainty reduction) in belief $p_{c}(V)$ induced by macro action $\lambda(v_i)$ is defined as:
\begin{align}
    I(V \, | \, \lambda(v_i)) &= \underbrace{H_{p_{c}}(V)}_\text{current entropy} - \underbrace{H_{p_{c}}(V^\prime \, | \, \lambda(v_i))}_\text{future entropy}
\end{align}
where the second term represents the expected future entropy of the graph after the agent has executed macro action $\lambda(v_i)$. 


\ph{Generalized Reward} To capture energy and distance measures in the coverage planning, we define the one-step reward to be the weighted sum of information gain and distance traveled under the macro action $\lambda(v_i)$:
\begin{align}
    Rew(v_i \, , \, \lambda(v_i)) &= w_1 \; I(V \, | \, \lambda(v_i)) - w_2 \; \ell_{e_{ij}}
\end{align}
where $w_1$ and $w_2$ weigh the information gain and traveled distance, respectively. 


\ph{Graph Policy Optimization}
Let $v(k)$ denote the $k$-th node visited under graph policy $\lambda$. Similarly, let $e(k,k+1)$ denote the edge between nodes $v(k)$ and $v(k+1)$.
We solve for the mapping $\lambda$ that maximizes the sum of future expected reward:
\begin{align}
    \lambda^* &= \arg\max_{\lambda(\cdot)} \, \mathbb{E} [ \sum_k Rew(v(k) \, , \, \lambda(v(k))) ]
\end{align}
This optimization can be solved via value iteration-based methods or forward search methods.

\begin{figure}[t!]
  \centering
  \includegraphics[width=0.4\textwidth]{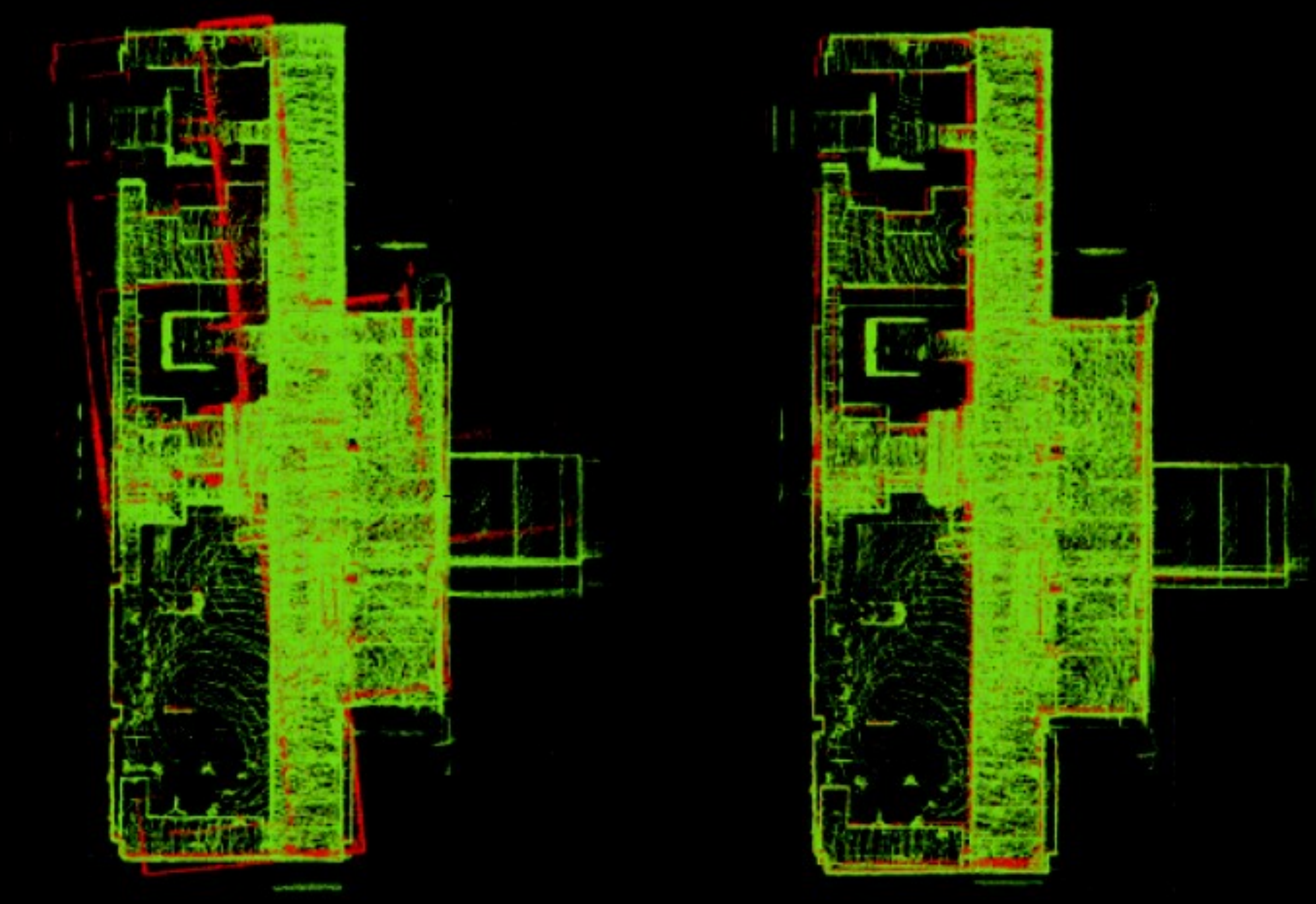}
  \caption{Map created by Au-Spot exploring Eagle Rock Substation, with different odometry sources: proposed method in green against KO in red (left) and KVO in red (right).}
  \label{spot_eagle_rock}
\end{figure}

\begin{figure}[t!]
  \centering
  \includegraphics[width=0.4\textwidth]{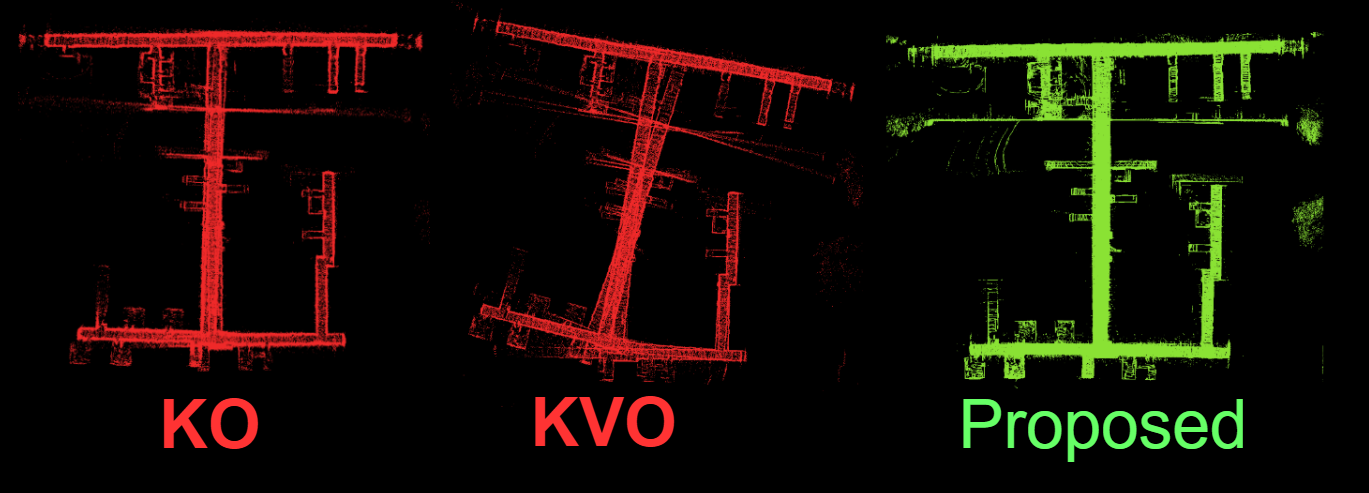}
  \caption{Map created by Au-Spot while exploring an office building at NASA's JPL with different odometry sources: KO (left), KVO (middle), and the Proposed method (right).}
  \label{spot_indoor_office}
\end{figure}

\section{Experimental Results}\label{sec:experiments}

The NeBula autonomy architecture is implemented on two Boston Dynamics Spot robots and  field-tested in subsurface, multi-level, and perceptually-degraded GPS-denied environments, including underground unstructured environments and industrial power plants.

\begin{figure}[t!]  
  \centering
  \includegraphics[width=0.45\textwidth]{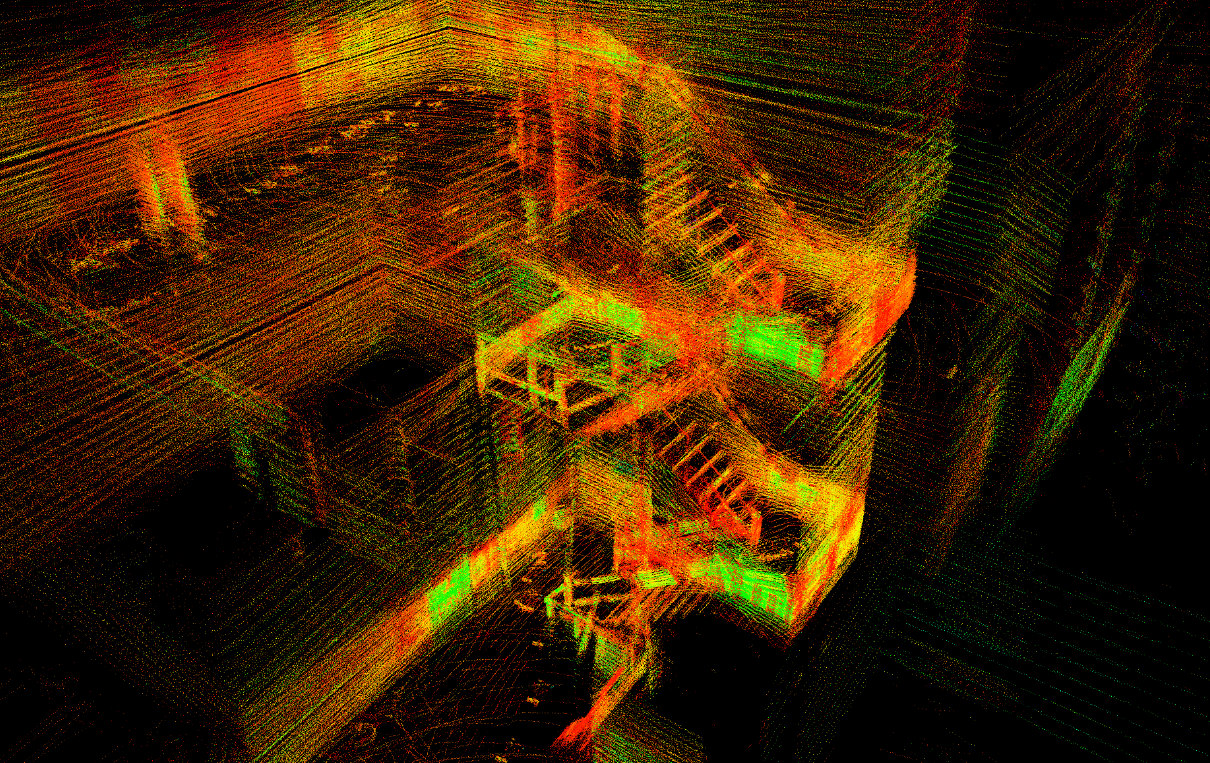}
  \caption{3D map generated by NeBula while traversing four flights of stairs. The map was created in real-time during the run (see Fig. \ref{fig:stairs-firstPage}).}
  \label{fig:alpha_course_stairs_map}
\end{figure}

As part of the Urban Circuit of the DARPA Subterranean Challenge, two Au-Spots were deployed into an Industrial Plant in February 2020 for live-mission exploration and  3D-Mapping of the environment (Fig. \ref{fig:alpha_course_stairs_map}).  The missions included detection of artifacts of interest such as backpacks, human survivors, gas-leaks, and cellphones via different sensing modalities including vision, thermal, gas sensors and wifi detection among others. The competition divided into four live-missions in unknown environments. 
Our Au-Spots explored a combined distance of 4km, including successfully climbing multiple flights of stairs. Points were awarded for accurate artifact detection and for successfully reporting this information back to the base-station.   
The NeBula framework successfully detected and localized a total of 16 artifacts, giving team CoSTAR a 1st-place win.



\ph{Odometry Estimation}
To demonstrate the performance of the proposed odometry pipeline, we compare the localization accuracy achievable using individual sensing channels with the proposed uncertainty-aware multi-sensor approach in perceptually-degraded environments. Figure \ref{spot_eagle_rock} depicts the results of the proposed method on data collected in the Eagle Rock subway station, Los Angeles. The rough and varying terrain causes KO to perform poorly, while a large amount of visual features causes KVO to produce a sufficiently accurate map. Conversely, Fig. \ref{spot_indoor_office} depicts results from data collected in NASA JPL's 198/161 offices. In this case, the soft carpet on the floor results in KO providing much more accurate maps than KVO.  KVO is instead challenged by feature-less white walls and the repetitive visual textures of the carpet. The different features of various environments make perception and odometry estimation challenging to a single sensing channel alone, hence the need for multi-channel odometry fusion. As seen in both figures, the proposed odometry generation method results in more accurate maps than those obtained by KO or KVO-based odometry.

\begin{figure}[t!]
  \centering
  \includegraphics[width=0.4\textwidth]{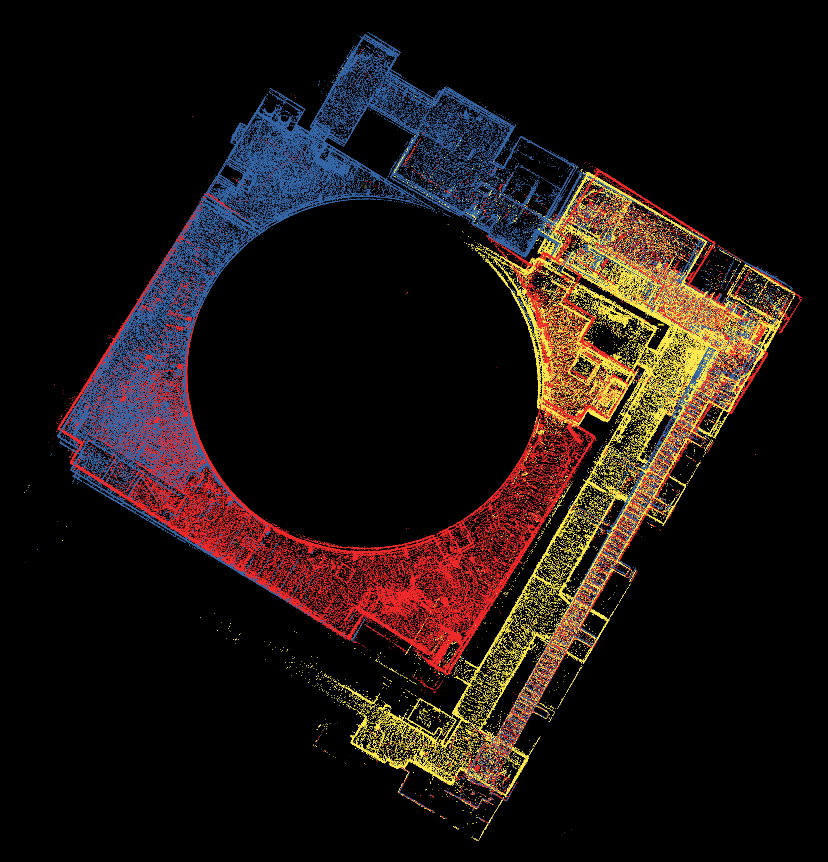}  
  \caption{Top view of area covered in urban environment by robot fleet consisting of one Au-Spot [yellow] and two UGVs [red/blue] during one run of the SubT Challenge, Urban Circuit. The map contains three different floors connected by multi-flight staircases. Note the complex topology (narrow passages, varying-sized interconnected rooms, and the outer circular geometry) requiring an irregular exploratory behavior that is characterized by a large looping path combined with many auxiliary paths necessary for the inspection of smaller structures.}
  \label{fig:exploration_planner_topview}
\end{figure}

\begin{figure}[t]
  \centering
  \includegraphics[width=\linewidth,trim={0cm 0cm 7cm 0cm},clip]{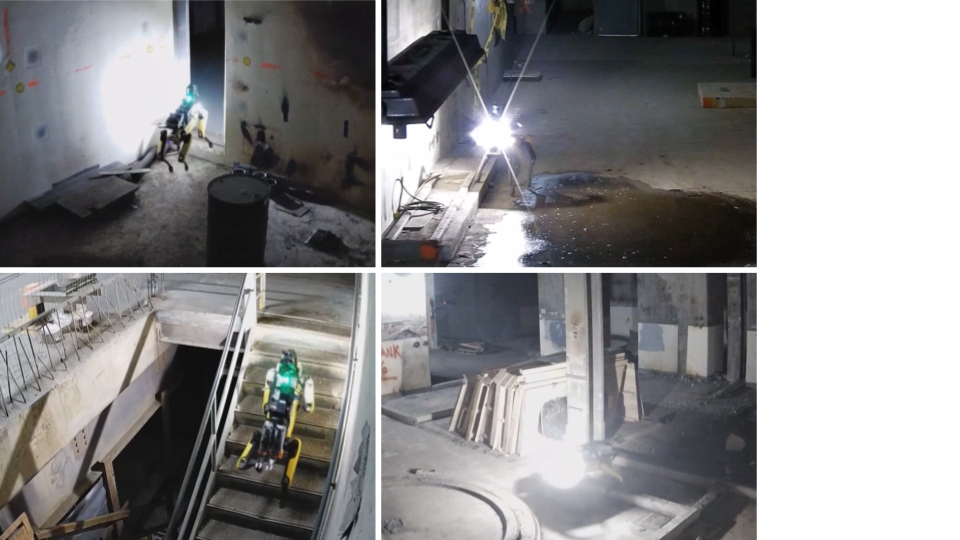}
  \caption{Narrow corridors (Top Left), water/oil (Top Right), stair wells (Bottom left), and raised concrete slabs (Bottom Right) are examples of some of the difficult terrain successfully navigated by Au-Spot. For full mission video, see \cite{spot_urban_video}.}
  \label{fig:trav_exp}
\end{figure}

\ph{Traversability}
Our perception-aware local planner enabled Au-Spot to safely and efficiently navigate through difficult environments, which contained a variety of unstructured obstacles and terrain, including ramps, slopes, piles of twisted metal, machinery and rubble (Fig. \ref{fig:trav_exp}).  Negative obstacles such as dropoffs and holes were also successfully avoided.

\ph{Coverage Planner}
Au-Spot's coverage planner successfully guided a fleet of two Au-Spots and two wheeled UGV robots to collaboratively explore and map a large unstructured environment within the one hour time limit in the SubT Challenge. Fig. \ref{fig:exploration_planner_topview} depicts the area explored by the robots during one of the four live-mission runs. In this single (one hour) run, the fleet of robots mapped a total volume of approximately 25,000 ${m}^{3}$. 
One of the most challenging elements of the course was traversing 4 flights of stairs, which induces pitching motions, physical slippage on the stair edges, and poor visual cues due to repetitive patterns of the stairs and railings. Fig. \ref{fig:stairs-firstPage} shows Au-Spot successfully climbing down stairs, and Fig. \ref{fig:alpha_course_stairs_map} depicts the map produced during the stair climbing operations, which allows the robot to accurately localize artifacts on multiple levels of the industrial power plant.

\ph{Complex Mission}
The video in \cite{spot_urban_video} depicts Au-Spot navigating and mapping one of the courses in the Urban Circuit of the DARPA SubT Challenge under time, communication, and computation constraints. The video begins as Au-Spot leaves the staging area where the human supervisor sends the ``go-command". Thereafter, Au-Spot's behavior at various phases of the mission, including when it is searching for artifacts, descending stairs, and deploying communication nodes, are presented. 
The video highlights how autonomous traversability and coverage planning enables the robot to thoroughly explore the environment, which includes challenging features such as hallways with narrow openings, large open rooms, raised platforms, and rubble. Over more than 1 km travel distance, the system was able to detect and localize artifacts while maintaining a localization error of less than 5~m -- the maximum artifact localization error to score points in the SubT Challenge.


\section{Conclusions}\label{sec:conclusions}
Motivated by exploring extreme environments and in particular underground environments in DARPA Subterranean Challenge, this system-focused paper discusses our developments toward endowing legged robots with hardware and perception capabilities required for high-levels of autonomy in extreme environments.
Specifically, we have presented our NeBula autonomy architecture applied to Boston Dynamics' Spot robot. NeBula is an architecture for risk- and perception-aware autonomy, applicable to a wide range of robots. In this paper, we have discussed a few representative NeBula modules, including odometry, traversability, and coverage planning, pertaining to the DARPA Subterranean Challenge. 
We believe this work takes an important step in advancing the state-of-the-practice and demonstrates the capabilities of legged robots for accomplishing complex, real-world, live-missions in extreme environments.

\section*{ACKNOWLEDGMENT}
The work is partially supported by the Jet Propulsion Laboratory, California Institute of Technology, under a contract with the National Aeronautics and Space Administration (80NM0018D0004), and Defense Advanced Research Projects Agency (DARPA).
We thank our team members, including
Angel Navarro,
Benjamin Morrell,
Hov Malekian,
Michael Wolf,
Fernando Chavez,
Alessandro Busichio,
Nobuhiro Funabiki,
Jeffrey Edlund,
Brett Lopez,
and
Kamak Ebadi
for their great contributions to this work.


\bibliographystyle{IEEEtran}
\bibliography{main}
\end{document}